\documentclass[lettersize,transaction]{IEEEtran}
\usepackage{amsmath,amsfonts}
\usepackage{algorithmic}
\usepackage{algorithm}
\usepackage{array}
\usepackage[caption=false,font=normalsize,labelfont=sf,textfont=sf]{subfig}
\usepackage{textcomp}
\usepackage{stfloats}
\usepackage{url}
\usepackage{verbatim}
\usepackage{graphicx}
\usepackage{cite}
\IEEEoverridecommandlockouts

\ifCLASSINFOpdf
\else
\fi
\begin{document}
%
\title{Visual Tactile Sensor Based Force Estimation \\ for Position-Force Teleoperation}
%
%

\author{Yaonan Zhu, Shukrullo Nazirjonov, Bingheng Jiang, Jacinto Colan, Tadayoshi Aoyama, Yasuhisa Hasegawa,\\
Boris Belousov, Kay Hansel, and Jan Peters
\thanks{This work is supported by JST AIP Grant Number JPMJCR20G8, Japan, and JSPS KAKENHI Grant Number JP22K14222.}
\thanks{Y. Zhu, S. Nazirjonov, B. Jiang, J. Colan, T.Aoyama, and Y. Hasegawa are with the Department of Micro-Nano Mechanical Science and Engineering, Nagoya University, Japan. \{zhu, shuk, bhjiang colan\}@robo.mein.nagoya-u.ac.jp, tadayoshi.aoyama@mae.nagoya-u.ac.jp, hasegawa@mein.nagoya-u.ac.jp. (Corresponding author: Yaonan Zhu)}
\thanks{B. Belousov, K. Hansel, and J. Peters are with the Department of Computer Science, Technical University of Darmstadt, Germany. \{boris.belousov, kay.hansel, jan.peters\}@tu-darmstadt.de.}
\thanks{This work has been submitted to the IEEE for possible publication.
Copyright may be transferred without notice, after which this version may
no longer be accessible.}
}

\markboth{Preprint}%
{Shell \MakeLowercase{\textit{et al.}}: Bare Demo of IEEEtran.cls for IEEE Journals}
%



\maketitle

\begin{abstract}
Vision-based tactile sensors have gained extensive attention in the robotics community. The sensors are highly expected to be capable of extracting contact information i.e. haptic information during in-hand manipulation. This nature of tactile sensors makes them a perfect match for haptic feedback applications. In this paper, we propose a contact force estimation method using the vision-based tactile sensor DIGIT \cite{Lambeta2020DIGIT}, and apply it to a position-force teleoperation architecture for force feedback. 
The force estimation is carried out by (1) building a depth map for DIGIT gel's surface deformation measurement, and (2) applying a regression algorithm on estimated depth data and ground truth force data to get the depth-force relationship. 
The experiment is performed by constructing a grasping force feedback system with a haptic device as a leader robot and a parallel robot gripper as a follower robot, where the DIGIT sensor is attached to the tip of the robot gripper to estimate the contact force.  
The preliminary results show the capability of using the low-cost vision-based sensor for force feedback applications.

\end{abstract}


%
\IEEEpeerreviewmaketitle

\section{Introduction}

Tactile sensors have gained extensive attention over the past years in the robotics community. Compared with conventional Force Torque sensors (FT sensors) which focus on precise measurement of contact forces, tactile sensors are inspired by human cutaneous perception and have advantages such as capturing surface deformation, detecting texture distribution, and detecting incipient slip \cite{dong2019maintaining}. Such multi-modal perception ability makes tactile sensors a suitable candidate to enhance the dexterity and performance of robotic hands \cite{fishel2012sensing}. 

Tactile sensor designs have several variations, each utilizing different sensing modalities. For example, BioTac® uses a pressure transducer to detect vibrations as small as a few nanometers for contact sensing \cite{reinecke2014experimental}, and the soft magnetic skin developed by Shen et al. detects normal and shear force through the change of magnetic flux densities \cite{yan2021soft}. On the other hand, vision-based tactile sensors provide cost-efficient but promising solutions for tactile sensing \cite{yuan2017gelsight}. In vision-based tactile sensors, contact information is extracted by performing computer vision algorithms on streamed image sequences from a camera (or cameras) mounted inside the sensors, which makes the sensor easy to access, and easy to implement \cite{lepora2022digitac}.

The applications of vision-based tactile sensors can be summarised as complementing robot tactile perception and enhancing the autonomous manipulation skills of robotic grippers to achieve human-like in-hand dexterity. 
In \cite{belousov2019building}, a library of tactile skills was developed by utilizing force, slip, and visual information from a Finger Vision sensor. Similarly, \cite{belousov2022robotic}, an end-to-end approach for tactile-based insertion with deep probabilistic ensembles and model predictive control was evaluated in contact-rich assembly tasks.
Moreover, Yuan et al. have successfully estimated object hardness through a sophisticated surface deformation analysis algorithm \cite{yuan2017shape}.

However, the full potential of the vision-based sensors should not be limited to autonomous in-hand robotic applications. As the design of tactile sensors is deeply influenced and inspired by the intrinsic human capability of haptic sensation, it makes a perfect match for haptic applications i.e. teleoperation with force feedback. 
Teleoperation with force feedback is also known as bilateral control systems, in which a leader device and a follower device are dynamically coupled to reflect physical interactions from the follower to the leader and vice versa \cite{anderson1989bilateral}. The research of bilateral control has a long history since the idea was first proposed by Goertz in the 1940s \cite{goertz1964manipulator}. Throughout the development of bilateral control, it has been experimentally applied to space robotics, telesurgery, and hazardous material handling \cite{hokayem2006bilateral}.

Bilateral control with force feedback is a direct control strategy that aims for ideal and seamless kinematic and dynamic coupling of leader and follower robots.
It can be categorized into two control architectures: (1) position-position architecture, and (2) position-force architecture \cite{siciliano2008springer}. 
In the simplest case, position-position architecture implements proportional-derivative (PD) controllers on both leader and follower sides and tracks each other. In this sense, users will feel the follower robot's driving forces as force feedback. However, the structure will present not only the surface contact force but also the inertia and other dynamic forces that drive the follower robots in free space. These extra forces' impact on the control becomes severe when the leader and follower have substantially different dynamic properties. 

To mitigate the undesired forces during free space movement, a position-force architecture is applied. In the control architecture, a force sensor is usually attached to the tip of the follower robot and it is used to measure contact force for feedback. 
The architecture only feedbacks the external forces acting between the follower robot and the environment and this makes the user's force perception clear. To enhance the stability of position-force architecture, feedback scaling, or passivity theory are used in general \cite{zhu2020enhancing}.  

In this paper, we adopt this position-force bilateral control for force feedback in robotic gripper teleoperation. The aim of this paper is 
to show the capability of vision-based tactile sensors on haptic (force) feedback applications. We use a vision-based tactile sensor for normal force estimation and integrate it with a position-force bilateral control system. 
The contributions of this paper are summarized as follows:
\begin{enumerate}
    \item A polynomial regression-based force estimation is performed to build a mapping of force value and measured depth value from vision-based tactile sensor DIGIT \cite{Lambeta2020DIGIT}. 
    \item Position-Force bilateral control that utilizes force estimation from DIGIT is implemented for force feedback in a teleoperated grasping system. 
    \item The force feedback performance of the proposed system is evaluated by conducting preliminary experiments on rigid object contact and a teleoperated in-hand pivoting task.
\end{enumerate}

\section{Position-Force Architecture and Force Estimation by Vision-Based Tactile Sensor}

In this section, the implemented position-force architecture is described and the details of the force estimation algorithm is given.

\subsection{Position-Force Architecture}
\begin{figure*}[t]
	\centering
	\includegraphics[width=0.9\linewidth]{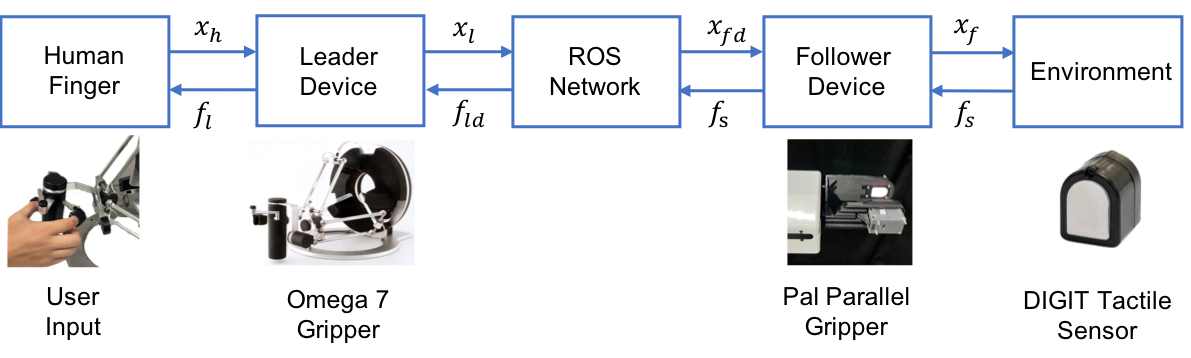}
	\caption{ Position-Force architecture and its integration with DIGIT tactile sensor.  }
	\label{fig:1}
\vspace{-5mm}
\end{figure*}

Fig. \ref{fig:1} shows the position-force architecture, in which tactile sensor DIGIT is used to substitute conventional FT sensor, and $x_h$, $x_l$, $x_{fd}$, $x_f$, $f_l$, $f_{ld}$, $f_{s}$, denotes for human command position, leader position, follower desired position, follower position, leader force, leader desired force, measured force from DIGIT sensor, respectively. Here the leader robot is the gripper of the Omega 7 haptic device (Force Dimension) and the follower robot is a parallel robotic gripper by PAL robotics.
The data communication is achieved through the ROS framework.
The control law of the position-force architecture is given as follows,
\begin{equation}
\begin{aligned}
   f_{ld} & = f_s, \\
   x_{fd} & = x_{l}
\end{aligned}
\end{equation}
where, the leader robot is force controlled to track measured force from the DIGIT sensor $f_s$, and the term $x_{fd}$ is an input to the position controller of the follower robotic gripper as a reference position to track. Through this position control, the gripping force can be modulated by teleoperation.

\subsection{Force Estimation by Polynomial Regression and Depth Map from DIGIT}
\begin{figure}[t]
	\centering
	\includegraphics[width=0.9\linewidth]{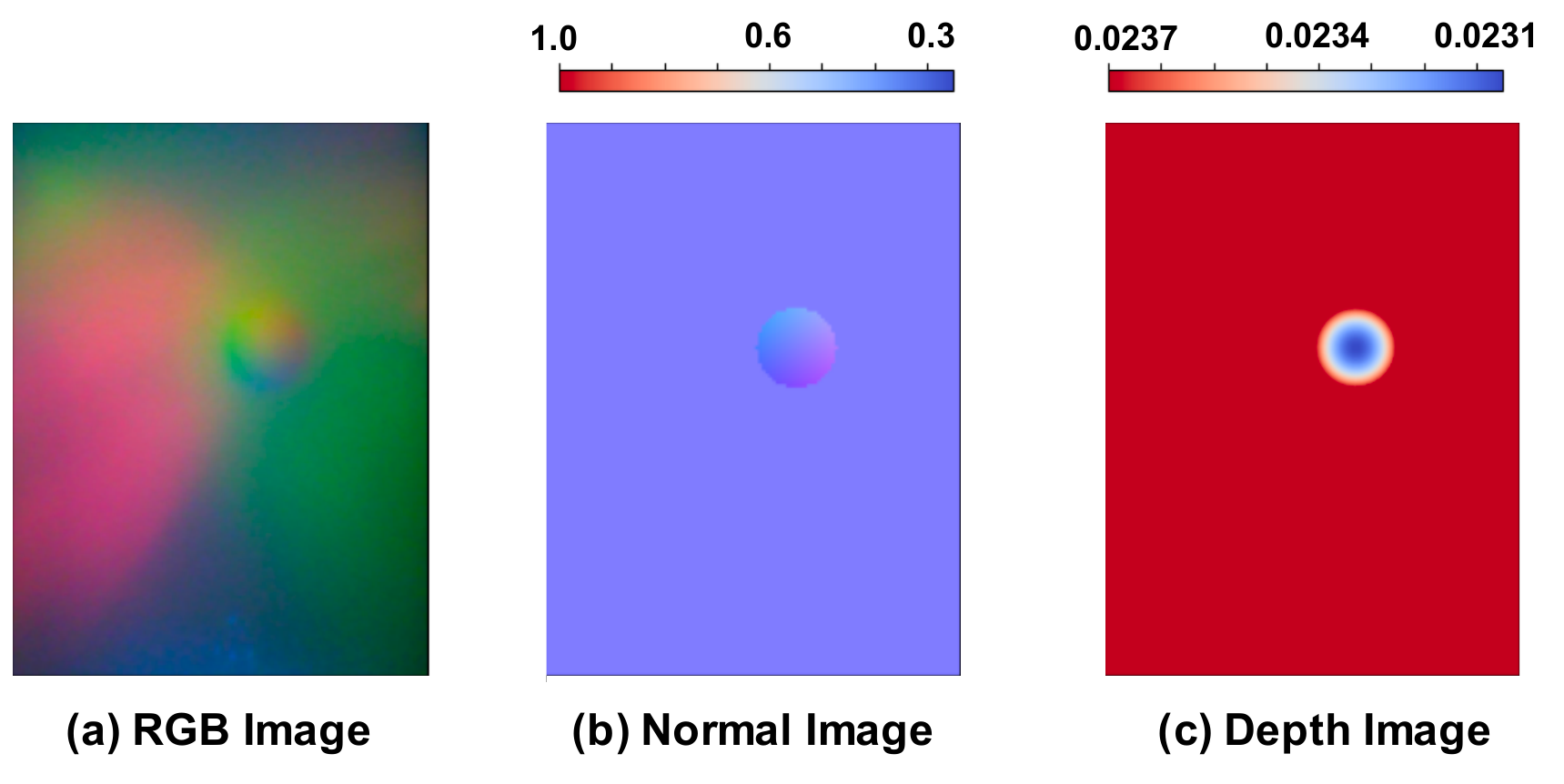}
	\caption{ Construction of a depth image from an RGB image. First, through MLP, an RGB image (a) is transformed into a normal image (b), then by calculating image gradients and applying a fast Poisson solver, we can construct a depth image.}
	\label{fig:2}
\end{figure}

This section explains the details of how the depth maps are extracted from DIGIT RGB images. 
The depth map is defined as the deformation level of each pixel on the gel surface.
The process is shown in Fig. \ref{fig:2}.
To get the depth map, we first train a three-layer Multi-Layer Perceptron (MLP) that learns a mapping from RGBXY values to surface normal values represented as $n_x$,$n_y$, and $n_z$, respectively. We used the Adam optimizer with a learning rate of 0.001 to minimize the MSELoss.
In addition, a dropout rate of 0.05 is included. 
We collect 40 RGB images by pressing a steel ball on the gel to train the MLP model. The steel ball has a diameter of 6 mm, and theoretically, the hemispherical shape enables us to extract surface normals in all directions \cite{yuan2017gelsight}. 
Surface normal image has a linear relationship with
the surface gradients as described by the following equations,
\begin{equation}
    G_x = \frac{n_x}{n_z}, G_y = \frac{n_y}{n_z}
\end{equation}
where, $G_x$, and $G_y$ represent image gradients in the X and Y directions, respectively. 
By applying this equation, we can obtain the surface gradients from surface normal images.
Given the surface gradients, by using a fast Poisson solver with Discrete Sine Transform (DST), we can finally build the depth map of the gel surface, similar to prior work by Wang et al. \cite{wang2021gelsight}, and Paloma et al \cite{sodhi2022patchgraph}. 

Once the depth map is generated, we select the maximum depth value from the depth map. This maximum depth value corresponds to the maximum surface deformation of the DIGIT gel during contact. Then, by building a regression model to map the gel deformation and ground truth contact force, we can later estimate the contact force in real time. This process is similar to using Young's modulus to estimate the counter force from an object with respect to surface deformation level.

\begin{figure}[t]
	\centering
	\includegraphics[width=\linewidth]{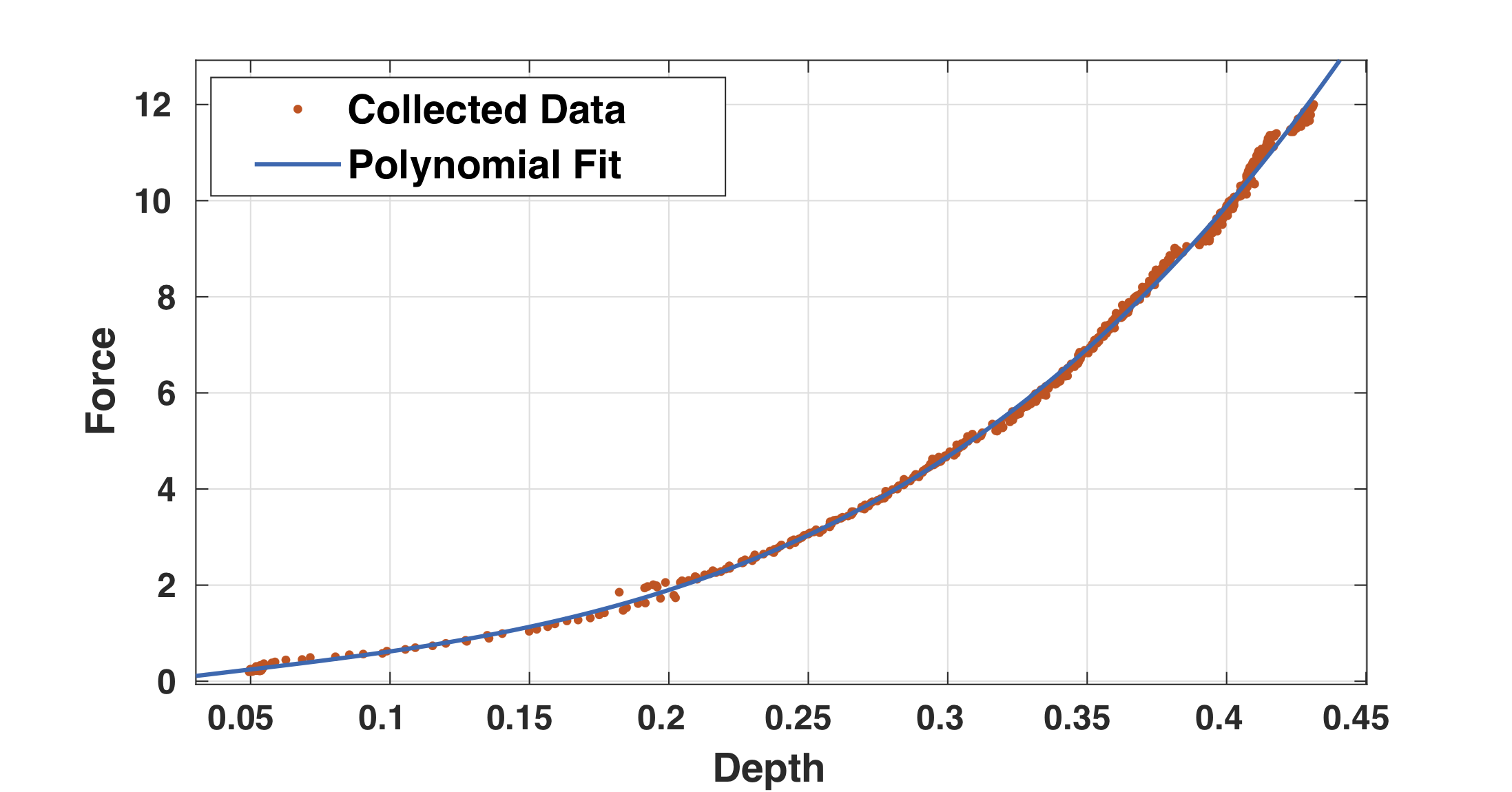}
	\caption{ Polynomial fit of collected data. }
	\label{fig:3}
\vspace{-4mm}
\end{figure}

To acquire the mapping between depth values and force values, we programmed the Omega 7 haptic device to press downwards (Z direction) with a constant displacement at each press. A rigid cylinder shape probe with a 5 mm radius and a curved surface facing down is attached to the tip of the device which is in direct contact with the DIGIT gel. The force value is read from the Omega 7 haptic device, and the depth value is read from the DIGIT sensor.
The orange dots in Fig. \ref{fig:3} show the collected force data with respect to depth value when pressing the DIGIT gel. 
Then we applied a least-square regression with 3 degrees polynomial to fit the data. The 3 degrees polynomial is defined as
\begin{equation}
    p(x) = p_1 x^3 + p_2 x^2 + p_3 x +p_4
\end{equation}
where $p_n$ denotes for coefficients. Once we get $p(x)$, we can use it for real-time force estimation. The result of polynomial fit is shown by a blue line in Fig. \ref{fig:3}, where the R-square measurement is 0.9987. 
The estimated force is then published as a ROS topic with a 30HZ refreshing rate which is the same as the camera's Frames Per Second (FPS).

\section{Experiments and Results}
In this section, we evaluate the preliminary performance of the force feedback with a vision-based tactile sensor by conducting rigid object contact and in-hand pivoting experiments.
\subsection{Rigid object contact and force feedback}
\begin{figure}[t]
	\centering
	\includegraphics[width=0.9\linewidth]{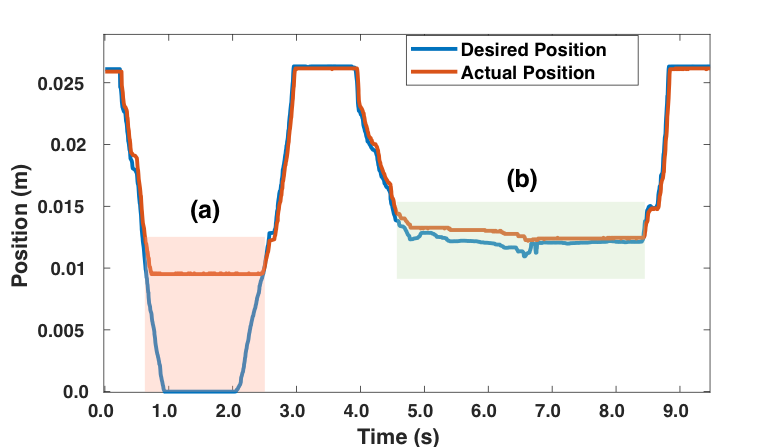}
	\caption{ Comparison of the desired position and actual position in a rigid object contact experiment. Region (a) represents without force feedback, and region (b) represents with force feedback}
	\label{fig:contact}
\vspace{-4mm}
\end{figure}

A rigid object contact experiment is conducted to show the force feedback capability. In the experiment, an operator controls the leader device to grasp a rigid object once without force feedback and once with force feedback. We treat the user's hand position $x_h$ as the desired position.
As Fig. \ref{fig:contact} shows, in the beginning, the operator moves freely, then in the region (a) the operator has contact with a rigid object without any force feedback, which leads to a large deviation from the desired position. In region (b), the operator contacts the rigid object with force feedback from the DIGIT sensor. 
The mean value of the desired position in region (b) is 0.0121 m, and the mean value of the actual position is 0.0127 m. Hence, the mean error is 0.0006 m.
The operator could maintain surface contact without large positional deviation and can feel the rigid object's location. In the free space movement, due to the nature of position-force architecture, no follower dynamics are felt by the operator.

\subsection{In-hand pivoting}
A preliminary experiment of teleoperated in-hand pivoting is carried out to evaluate the performance of the proposed force feedback architecture. 
In-hand pivoting is an important dexterous manipulation skill in robotics.
Through in-hand pivoting, a robotic grasping system can perform repositioning tasks and hence compensate for environmental uncertainties and imprecise motion execution \cite{8202299}.
Humans rely on haptic sensations to perform this dexterous task, and in this preliminary experiment, we evaluate the haptic feedback performance of the proposed architecture for such a teleoperated dexterous task.

\begin{figure}[t]
	\centering
	\includegraphics[width=\linewidth]{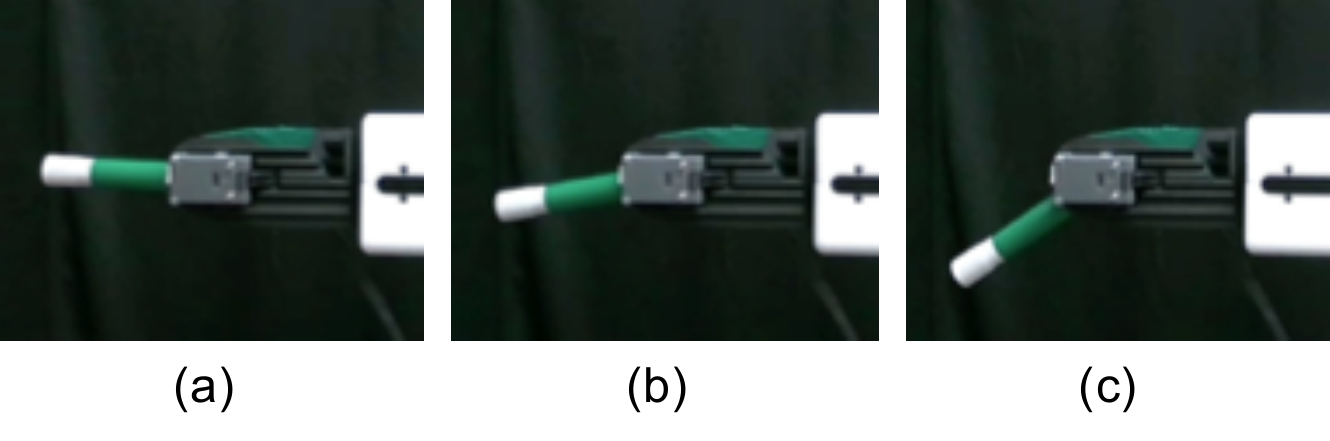}
	\caption{ A successful pivoting of a cylinder marker to target orientation.}
	\label{fig:pivot}
\vspace{-5mm}
\end{figure}

In this experiment, as Fig. \ref{fig:pivot} shows, the experiment task is to pivot a grasped cylinder marker from a horizontal position (Fig. \ref{fig:pivot}a) to align with the bottom left corner of the DIGIT sensor (Fig. \ref{fig:pivot}c). The surface friction can be controlled by controlling the distance between the gripper fingers, i.e. wider finger distance reduces friction hence the object will pivot because of gravity. Throughout the experiment, visual feedback with a fixed point of view is given (Fig. \ref{fig:pivot}). 
We evaluated two conditions, Visual + Force feedback and Visual feedback.
Three subjects participated in the experiment. During the experiment, subjects pivot the object 5 times for each condition. Task completion time and success rate are measured.

\begin{figure}[t]
	\centering
	\includegraphics[width=0.85\linewidth]{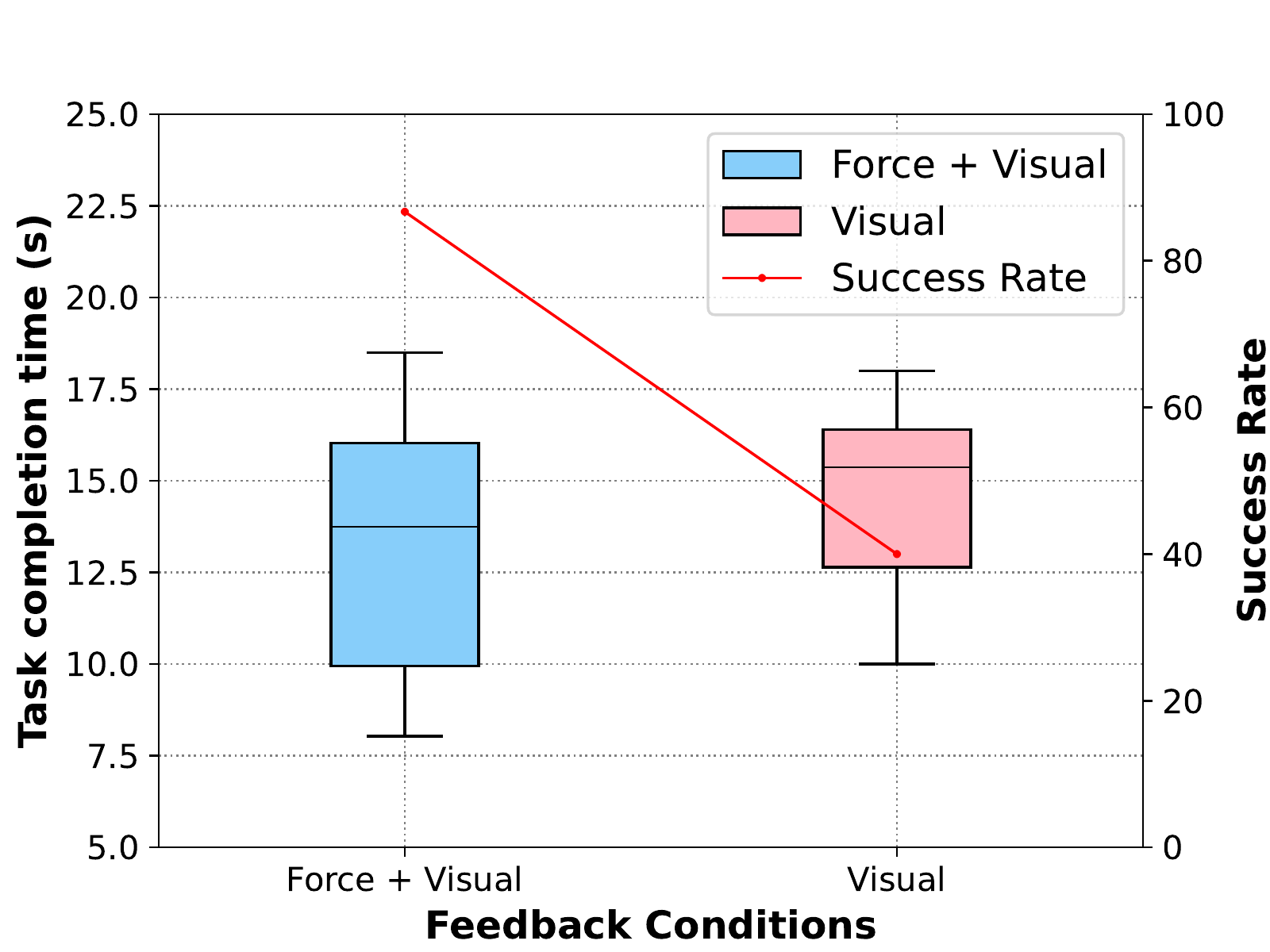}
	\caption{ Box plot of pivoting completion time and task success rate for two feedback conditions. The red line represents the task success rate.}
	\label{fig:pivot_exp}
\end{figure}

The experimental result is shown in Fig. \ref{fig:pivot_exp}. The blue and red plot represents the box chart of task completion time for the Visual + Force feedback condition and the Visual feedback condition, respectively. The median value of the visual + force feedback is 13.75, and the visual feedback condition is 15.37. The task success rate is 86.7\% and 40.0\%, respectively. 
The experimental results indicate that by using force feedback, subjects complete the pivoting with a higher success rate and slightly faster median completion time. 
However, in visual + force feedback, some trials required a higher completion time. This may be due to the subjects needing more time to adjust the grasping according to the perceived force, which requires further analysis in the future.  
The results also imply that the proposed position-force system can display the force feedback for operators, and proves the potential of tactile sensor DIGIT for such a force feedback application. 

\section{Conclusions and Future Works}
Vision-based tactile sensors that aim to enhance the dexterity of robotic manipulation systems have gained extensive attention in the community.  
While the nature of the tactile sensors makes them a good candidate for in-hand manipulation tasks, the sensors can potentially be applied to haptic feedback in teleoperation.
This paper proposes a force feedback methodology by integrating a position-force architecture and a vision-based tactile sensor DIGIT for contact force measurement. 
The force measurement is performed by learning a depth estimation model, followed by building a polynomial regression model that maps the relationship between force and depth.
In a rigid object contact experiment, the operator can feel the force feedback and maintain surface contact. In the preliminary experiment of in-hand pivoting, by using force feedback the subjects are able to complete the pivoting task with a faster median time and a higher success rate.  
The preliminary experimental results indicate that the vision-based tactile sensor DIGIT can be integrated with a teleoperation system to provide force feedback.
It would be interesting to further enhance the sensor performance with multi-point contact in future work. 
Moreover, the integration of haptic feedback and other sensing modalities such as shear force estimation, as well as surface slip prediction \cite{veiga2018grip} can be explored in the future.
\bibliographystyle{IEEEtran}
\bibliography{shared_autonomy}
%



%




\end{document}